# MTD: Multi-Timestep Detector for Delayed Streaming Perception

Yihui Huang[1] , Ningjiang Chen[1,2,3(✉)]

[1] School of Computer and Electronic Information, Guangxi University, Nanning , China
[2] Guangxi Intelligent Digital Services Research Center of Engineering Technology, Nanning, China
[3] Key Laboratory of Parallel, Distributed and Intelligent Computing (Guangxi University), Education Department of Guangxi Zhuang Autonomous Region, Nanning, China
`yulin_work@qq.com,chnj@gxu.edu.cn`

**Abstract.** Autonomous driving systems require real-time environmental perception to ensure user safety and experience. Streaming perception is a task of reporting the current state of the world, which is used to evaluate the delay and accuracy of autonomous driving systems. In real-world applications, factors such as hardware limitations and high temperatures inevitably cause delays in autonomous driving systems, resulting in the offset between the model output and the world state. In order to solve this problem, this paper propose the Multi-Timestep Detector (MTD), an end-to-end detector which uses dynamic routing for multi-branch future prediction, giving model the ability to resist delay fluctuations. A Delay Analysis Module (DAM) is proposed to optimize the existing delay sensing method, continuously monitoring the model inference stack and calculating the delay trend. Moreover, a novel Timestep Branch Module (TBM) is constructed, which includes static flow and adaptive flow to adaptively predict specific timesteps according to the delay trend. The proposed method has been evaluated on the Argoverse-HD dataset, and the experimental results show that it has achieved state-of-the-art performance across various delay settings.

**Keywords:** Streaming Perception, Object Detection, Dynamic Network

## 1    Introduction

To ensure user safety and experience, autonomous driving systems need to perceive surrounding environment not only in time, but also in real time. Traditional object detection[1,2,3] benchmarks focus on offline evaluation, which means that each frame of the video stream is compared with its annotation separately, which requires the system to process the captured frames within the interval time between each frame. Therefore, many researches[4,5,6,7,8] focus on reducing the latency so that the model can finish processing captured frame before next frame is input. However, in real-world applications, hardware limitations[9,10] and high temperatures[11] inevitably cause processing delay, resulting in changes in the real-time environment after model processes the captured frames. As shown in Fig. 1 (a), the output of the model is always outdated.

To solve the above problem, exsiting works use methods such as Optical Flow[12,13], Long Short-Term Memory (LSTM)[14,15] and ConvLSTM[16,17,18] to extract features and predict future frames. There are also works that combine video object tracking and detection[19,20] to extract the spatio-temporal information of the detected object, thereby optimizing the detection accuracy. However, these methods are typically designed for offline environments and do not account for latency in streaming perception scenarios.

Li et al.[21] point out that the difference between standard offline evaluation and real-time applications: the surrounding world may have changed by the time an algorithm finishes processing a frame. To address the problem, 'Streaming Average Precision (sAP)' is introduced, which integrates latency and accuracy to evaluate real-time online perception. This metric evaluates the entire perception stack in real-time, forcing model to predict current world state, as shown in Fig. 1(b).

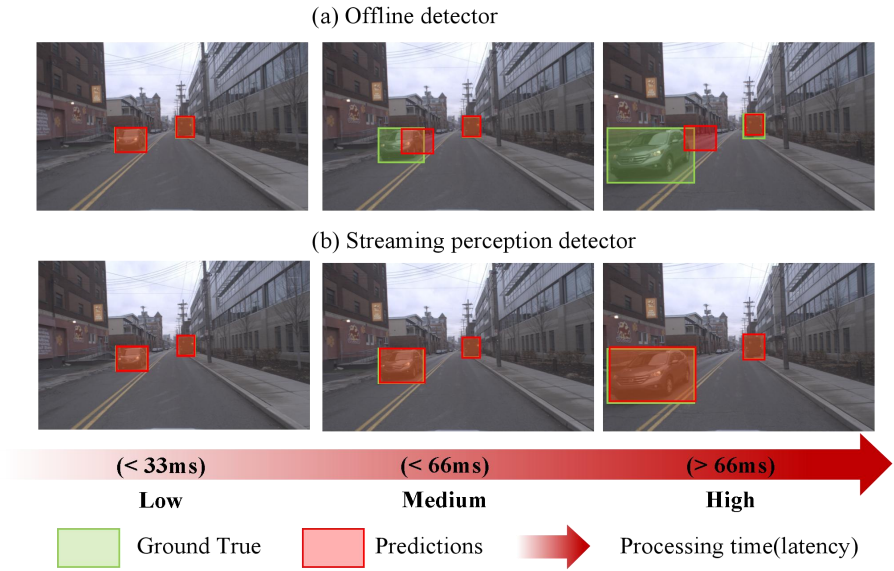

**Fig. 1.** Visualization of theoretical output of offline detectors vs. streaming perception detectors under various latency environments. Offline detectors (a) and streaming perception detectors (b) produce very different predictions when encountering processing delays.

StreamYOLO[22] suggests that the performance gap in streaming perception tasks is due to the current processing frame not matching the next frame, so they simplified streaming perception to the task of predicting the next frame. The work introduces Dual-Flow Perception (DFP) Module to merge the features of the previous and current frames, and Trend-Aware Loss (TAL) to adjust adaptive weights for each object based on its motion trend, achieving state-of-the-art sAP performance.

However, It can only predict the next frame, and its performance is optimal when the processing time is less than inter-frame time.

Based on the work of StreamYOLO, DaDe[23] suggested that automatic driving system should consider additional delays. It proposes Feature Queue Module to store historical features, and Feature Selection Module supervises processing delay. Based on the processing delay, the Feature Select Module adaptively selects different historical features for fusion to create object motion trends, predicting future with different timesteps. The study improves the baseline to handle unexpected delays, achieving the highest sAP performance with one additional timestep, but performance degradation increases when more timesteps are considered.

Although the above methods improve performance to some extent, the predicted timestep is still inadequate to keep up with the real-time environment when processing delay is substantial.

Dynamic networks[24] present a novel option for solving this problem. Unlike static models with fixed calculation graphs and parameters, dynamic networks can dynamically adjust structure during inference, providing properties that static models lack. Common dynamic networks include dynamic depth networks that distinguish sample difficulty by early exit[25,26] or skipping layers[27,28], and dynamic width networks[29,30] that selectively activate neurons and branches according to input during the inference process. Both methods adjust depth or width of the architecture by activating computing units to reduce model costs. In contrast, dynamic routing networks[31,32] perform dynamic routing in a supernetwork with multiple possible paths, by assigning features to different paths, allowing the computational graph to adapt to each sample during inference.

In this paper, a Multi-Timestep Detector (MTD) is proposed. The main idea is to transform the detection problem of delayed streaming perception into a future prediction problem with multiple timesteps. A Timestep Branch Module (TBM) is constructed to investigate the generality of backbone features on similar tasks. By redesigning the training pipeline, multiple detection head branches are trained to predict different timesteps in the future. The Delay Analysis Module (DAM) continuously monitors preprocessing and inference delays, calculates the delay trend, and analyzes the target timestep, enabling the model to choose the best branch for detection. Experiments are conducted on the Argoverse-HD[21] dataset, and MTD demonstrates significant performance improvements over various delay settings compared to previous methods. The contributions of the work are as follows:

- By analyzing the processing time fluctuations of the model, the inference delay trend is introduced for more reliable calculation of the delay trend. Based on the idea of dynamic routing, Timestep Branch Module is proposed, where the delay trend is used as the routing basis. Under various delay settings, these methods can effectively improve the accuracy of streaming perception.
- To the best of our knowledge, MTD is the first end-to-end model that uses dynamic routing for delayed streaming perception. Under various delay settings, our method improves the sAP[21] performance by 1.1-2.9 on the Argoverse-HD dataset compared with baselines.

## 2 The Methods

This section describes how MTD is designed to predict future at different timesteps. The Delay Analysis Module is constructed to calculate delay trend. The Timestep Branch Module dynamically routes within the network based on delay trend and produces outputs that align with the real environment. The training pipeline is designed to avoid adding extra computations during inference and maintain real-time performance. Fig. 2 illustrates the model's inference process.

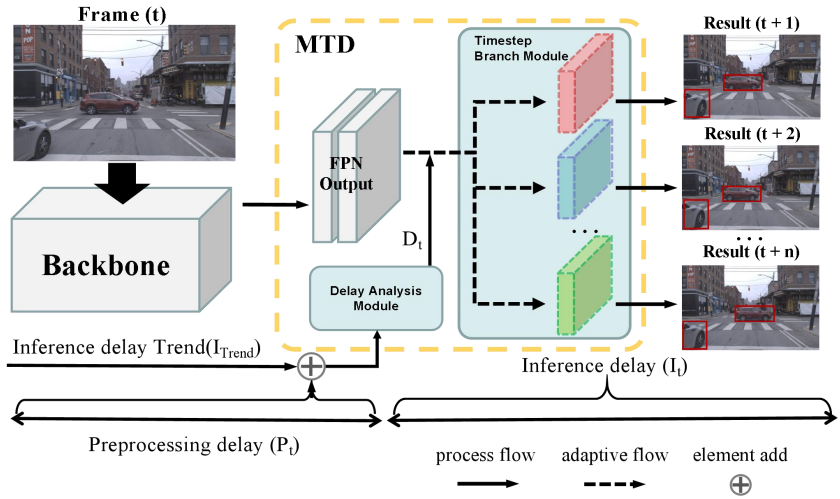

**Fig. 2.** MTD extracts features using a shared backbone. Delay Analysis Module (DAM) calculates delay trend $D_t$ according to current preprocessing delay $P_t$ and inference delay trend $I_{Trend}$. Timestep Branch Module (TBM) selects detection head adaptively to predict future at different timestep based on input $D_t$. The inference path of the detector is divided into process flow and adaptive flow, in which the process flow is a fixed inference path, and the adaptive flow is an optional path, which is dynamically routed by the TBM according to the delay trend.

### 2.1 Delay Analysis Module (DAM)

The Delay Analysis Module calculates delay trend for subsequent modules inference. In the work of DaDe[23], the delay trend $D_t$ of the current frame t is simply expressed as the sum of the preprocessing delay $P_t$ and the last frame inference delay $I_{t-1}$. This approach is simple and effective, but performs best only when the model's processing time trend is flat.

Fig. 3 visualizes frame processing time under none delay setting. It is observed that the model's processing time is mostly smooth, with occasional small peaks. As shown in Fig. 3 (a)-(c), model delay fluctuations often occur within a single frame, leading to the delay analysis method in [23] incorrectly assessing the delay after each anomaly frame. Consequently, subsequent modules struggle to hit the correct timestep. To solve this issue, Inference delay trend $I_{Trend}$ is introduced to reduce the sensitivity of inference delay perception. The formula for calculating $I_{Trend}$ is as follows:

$$I_{Trend} = \begin{cases} I_{t-1}, \frac{I_{t-1}}{I_{t-2}} < \tau \\ I_{t-2}, \frac{I_{t-1}}{I_{t-2}} \geq \tau \end{cases} \tag{1}$$

where the inference delay trend $I_{Trend}$ is the ratio of the inference delay from the previous two frames, $I_{t-1}$ and $I_{t-2}$, which evaluates the fluctuation of the processing delay. Hyperparameter $\tau$ a is obtained by grid search.

By calculating inference delay trend and preprocessing delay of the current frame, the delay trend $D_t$ is defined as follows:

$$D_t = P_t + I_{Trend} \tag{2}$$

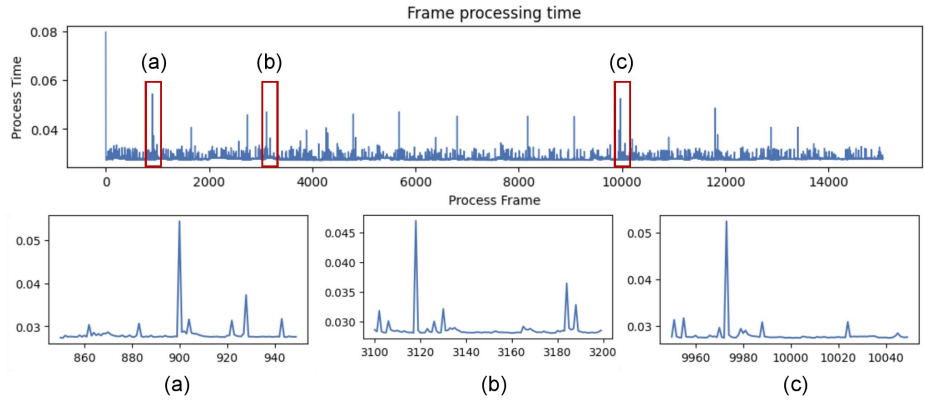

Fig. 3. Visualization of model frame processing time. The processing time of each frame is recorded in an environment without additional delay.

## 2.2 Timestep Branch Module (TBM)

The feature extracted by the backbone has generalization capabilities for similar tasks. Specifically, the feature contains moving trend and basic semantic information that can be used not only to predict the next frame but also to predict future frames at farther timesteps.

To this end, Timestep Branch Module (TBM) is designed for multi-branch decoding of features. To demonstrate the effectiveness of this method, StreamYOLO's detection head [22] was repurposed as a template for multiple branches, rather than designing a more complex one.

In StreamYOLO, the training input is a triplet consisting of previous frame $F_{t-1}$, current frame $F_t$, and the ground truth $F_{t+1}$. So the training dataset is reconstructed as $\{(F_{t-1}, F_t, F_{t+1})\}_{t=1}^N$, where $N$ is the total number of samples. To adapt this method for multi-branch structure model training, future frames with various timesteps is used as the predicted ground truth for different detection heads. Dataset is reconstructed as $\{\{(F_{t-1}, F_t, F_{t+s})\}_{t=1}^N\}_{i=2}^S$, where $S$ is the total number of detection heads and $i$ is the index of detection heads. The pre-trained head from StreamYOLO is skipped and started the index of $i$ from 2. During training process, the model trains each newly added detection head iteratively based on dataset. To prevent other parts of the model from being affected by backpropagation, the backbone and other detection heads are frozen during training. Fig. 4 visualizes the training process.

During the inference process, TBM receives delay trend $D_t$ from DAM as input , and calculates the target timestep $n$ according to delay trend $D_t$ and inter-frame time $T$ . The formula for target timestep $n$ is as follows :

$$n = \left\lfloor \frac{D_t}{T} \right\rfloor \tag{3}$$

Once $n$ is determined, TBM selects the optimal detection head $S_n$ using Formula 4. If the target timestep exceeds the maximum timestep, the detection head closest to the target timestep will be selected.

$$TBM([S_1, S_{max}], D_t) = S_n \tag{4}$$

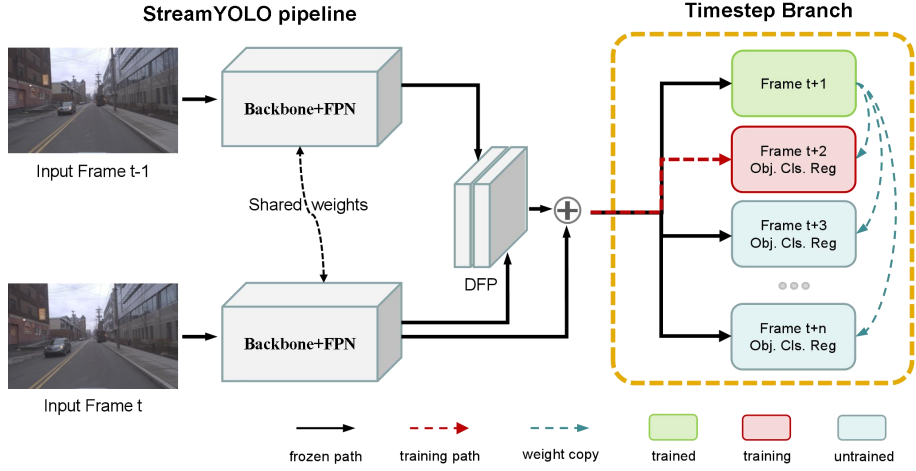

**Fig. 4.** Training process of MTD. To initialize all detection heads of the TBM, the detection head from StreamYOLO is used as a template and set the same weight. The training process begins with the t + 2 branch and gradually traverses all subsequent heads.

## 3 Experiments and Evaluation

### 3.1 Experimental Settings

*Dataset.*

Argoverse-HD[21] dataset provides high-resolution and high-frame-rate driving video image data, and provides sufficient streaming perception annotation data to evaluate the real-time performance of object detection algorithm. The train/val split in [21,22,23] is followed for evaluation and comparison with other relevant studies.

*Evaluation metrics.*

**Streaming Average Precision(sAP).** sAP[21] is a metric for streaming perception. It takes into account both latency and accuracy, and evaluates the model output under the framework of current world state. Similar to MSCOCO[33], The sAP toolkit[34] evaluates average mAP over Intersection-over-Union (IoU) thresholds from 0.5 to 0.95, and $sAP_S$, $sAP_M$, $sAP_L$ for different object size.

**Missed Timestep.** To assess the performance of DAM, a new metric called Missed Timestep is introduced. The delay trend can be obtained from Formula 2, while the actual delay trend is defined as Formula 5, which is the sum of the preprocessing time and inference time of the current frame.

$$AD_t = P_t + I_t \qquad (5)$$

By applying these formulas to Formula 3, the timestep *n* of the delay trend $D_t$ and timestep *m* of the actual delay trend $AD_t$ can be obtained. When *n* and *m* are equal, the model selected the best detection head. Conversely, when *n* and *m* are not equal, the model missed the timestep and selected a sub-optimal detection head. The lower the total number of Missed Timesteps, the better the performance of DAM.

*Implementation details.*

MTD is fine-tuned based on StreamYOLO[22] pre-training weight. Two additional detection heads are trained to cover common delays. The model is trained on four 1080Ti GPUs, and each detection head is trained for 5 epochs. Batchsize is set to 32. The model performance is evaluated on a 3090Ti GPU.

*Delay settings.*

Delay settings (none, low, medium, and high) are simulated by manually injecting processing delays. The medium delay simulated the fluctuation of delay across different timesteps to assess the model's performance in cases of severe delay fluctuations. Fig. 5 illustrates the processing time distribution under each setting. Inter-frame time is 33ms at 30FPS, so processing time should be less than 33ms to maintain real-time performance. As shown in Table 1, under none delay settings, all three models can maintain real-time performance. As the delay increases, evaluation requires the use of t + n frames that correspond to the real environment as the ground truth for comparison, and the model's prediction difficulty gradually increases.

**Table 1.** Mean delay and standard deviation in various delay settings. All delays are in milliseconds.

| Environment | None | | Low | | Medium | | High | |
|---|---|---|---|---|---|---|---|---|
| StreamYOLO | 27.8 | 1.63 | 59.2 | 2.82 | 69.6 | 4.31 | 90.4 | 3.47 |
| DaDe | 27.9 | 1.89 | 59.3 | 2.47 | 69.5 | 4.52 | 90.3 | 3.67 |
| Ours | 28.1 | 2.1 | 59.5 | 3.24 | 68.7 | 4.46 | 89.8 | 3.23 |

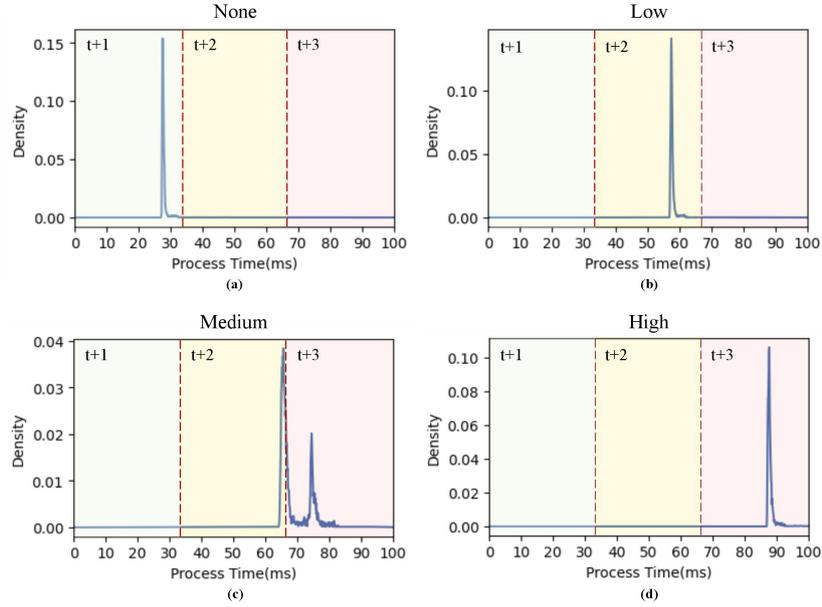

**Fig. 5.** Visualization of delay distribution for various delay environments. Vertical dotted line separates the inter-frame space.

### 3.2 Experimental Results

**Table 2.** Grid search for $\tau$ in Formula 1

| $\tau$ | None | Low | Medium | High |
|---|---|---|---|---|
| 0.70 | 72 | 12 | 1434 | 2 |
| 0.80 | 57 | 10 | 1369 | 2 |
| 0.90 | 46 | 10 | 1278 | 2 |
| . . . | | | | |
| 0.99 | 45 | 10 | 1166 | 2 |
| **1.00** | **45** | **10** | **1162** | **2** |
| 1.01 | 45 | 10 | 1164 | 2 |
| . . . | | | | |
| 1.10 | 47 | 10 | 1246 | 2 |
| 1.20 | 48 | 9 | 1306 | 2 |
| 1.30 | 54 | 10 | 1357 | 2 |

*Evaluation of delay analysis methods.*

The Delay Analysis Module calculates delay trends for subsequent inference modules according to inference time and preprocessing time. Incorrect delay trends can cause the Timestep Branch Module to hit the wrong timestep, resulting in poor inference performance. The performance of the Delay Analysis Module is assessed by counting the number of missed timesteps. According to the definition of Formula 1, the value of τ is used as a threshold to control the model's sensitivity to delay fluctuations. A grid search is performed on the value of τ, and the results are shown in Table 2. The Delay Analysis Module has the least number of missed timesteps when the value of τ is 1.00. Table 3 show the comparisons of our delay analysis method and DaDe's method in terms of missed timesteps under various delay settings. Our method reduces the number of missed timesteps by 14.4% to 30% compared to DaDe[23].

**Table 3.** Performance comparisons of delay analysis method (MTD vs DaDe).

| Environment | Processed Frame | Missed Timestep | Decrease Percent |
|---|---|---|---|
| None (27ms±1) | | | |
| DaDe | 15035/15062 | 58 | 22.4% |
| MTD(τ=1) | | 45 | |
| Low (59ms±1) | | | |
| DaDe | 8285/15062 | 13 | 23.0% |
| MTD(τ=1) | | 10 | |
| Medium (69ms±1) | | | |
| DaDe | 7164/15062 | 1359 | 14.4% |
| MTD(τ=1) | | 1162 | |
| High (90ms±1) | | | |
| DaDe | 5580/15062 | 3 | 30.0% |
| MTD(τ=1) | | 2 | |

*Comparison of computations.*

**Table 4.** Comparison of model parameters, Gflops and processing time.

| Environment | Params | Gflops | Processing Time (ms) |
|---|---|---|---|
| StreamYOLO | 54.84M | 383.47 | 27.84 |
| DaDe | 54.84M | 383.47 | 27.92 |
| MTD | 69.95M(+15.11) | 383.47 | 28.1(+0.26) |

The additional detection head increases the total parameters of the MTD, but the increased parameters are distributed on different paths, so the processing time and Gflops during inference are unchanged (see Table 4), which makes the processing time gap between the MTD and the baselines almost negligible. (27.84ms vs 28.1ms)

*Comparisons with SOTA Methods.*

A performance comparison is performed between MTD and two baselines (DaDe[23] and StreamYOLO[22]) under various delay settings, with the outcomes detailed in Table 5. Without extra delay, the performance of all detectors is similar. Under low delay, StreamYOLO's performance starts to lag behind the others as its frame processing time exceeds the frame interval time(33ms). As the processing delay increases, StreamYOLO cannot predict the future of farther timesteps, resulting in a widening performance gap. DaDe constructs long-term motion trends by fusing historical frame features with longer time intervals, resulting in a slower decline in performance under low-latency settings. However, as the delay trend moves towards medium and high delay, its performance declines more and more faster, even lower than StreamYOLO under high delay. This issue arises from the fused features being temporally distant, causing feature-level fusion ineffective in creating a motion trend, hence reducing the model's inference performance. Under different delay settings, our method improves sAP performance by 1.1-2.9, achieving the best performance across all metrics, further validating the effectiveness of our approach.

**Table 5.** Performance comparison with baselines in Argoverse-HD dataset.

| Environment | sAP | sAP$_{50}$ | sAP$_{75}$ | sAP$_S$ | sAP$_M$ | sAP$_L$ |
|---|---|---|---|---|---|---|
| None (27ms±1) | | | | | | |
| StreamYOLO | 36.9 | 58.1 | 37.6 | 14.7 | 37.4 | 64.2 |
| DaDe | 36.9 | 58.0 | 37.6 | 14.6 | 37.4 | 64.4 |
| MTD | 36.9 | 58.1 | 37.7 | 15.0 | 37.0 | 64.5 |
| Low (59ms±1) | | | | | | |
| StreamYOLO | 26.2 | 48.0 | 24.4 | 8.8 | 24.5 | 41.5 |
| DaDe | 27.8 | 49.0 | 26.8 | 9.8 | 27.2 | 42.3 |
| **MTD** | **29.1** | **51.5** | **28.0** | **10.2** | **28.3** | **46.6** |
| Medium (69ms±1) | | | | | | |
| StreamYOLO | 25.3 | 46.8 | 23.1 | 8.2 | 23.4 | 39.9 |
| DaDe | 25.3 | 46.9 | 23.5 | 8.2 | 23.6 | 39.7 |
| **MTD** | **26.4** | **49.0** | **24.2** | **8.7** | **24.7** | **43.1** |
| High (90ms±1) | | | | | | |
| StreamYOLO | 22.7 | 43.3 | 20.5 | 6.9 | 20.7 | 35.8 |
| DaDe | 22.4 | 42.2 | 20.1 | 7.0 | 20.5 | 34.5 |
| **MTD** | **24.1** | **46.1** | **21.4** | **7.9** | **21.6** | **39.1** |

# 4 CONCLUSION

This paper presents a Multi-timestep Detector[1] (MTD) that includes a Delay Analysis Module (DAM) to monitor the model's delay trend in real-time, and a Timestep Branch Module (TBM) that employs dynamic routing in multiple branches for multi-timestep prediction. The experimental results shows that MTD is particularly suitable for delayed streaming perception scenarios and achieves state-of-the-art performance under various delay settings. Moreover, the TBM demonstrates that the features extracted by the backbone are universally applicable to predicting future video frames with various timesteps. In the future, it is plan to investigate whether this method has generalization significance in other related tasks, such as video sequence generation and human pose prediction.

---

[1] The code is available at: https://github.com/Yulin1004/MTD.

# 5 ACKNOWLEDGMENT

This work is funded by the Natural Science Foundation of China (No. 62162003).

# 6 REFERENCES

1. Girshick, R., Donahue, J., Darrell, T., Malik, J.: Rich feature hierarchies for accurate object detection and semantic segmentation. In: IEEE Conference on Computer Vision and Pattern Recognition. pp. 580–587 (2014).
2. Girshick, R.: Fast R-CNN. In: IEEE International Conference on Computer Vision. pp. 1440-1448 (2015).
3. Ren, S., He, K., Girshick, R., Sun, J.: Faster R-CNN: Towards Real-Time Object Detection with Region Proposal Networks. Advances in neural information processing systems 28 (2015).
4. Redmon, J., Divvala, S., Girshick, R., Farhadi, A.: You Only Look Once: Unified, Real-Time Object Detection. In: IEEE Conference on Computer Vision and Pattern Recognition. pp. 779-788 (2016).
5. Redmon, J., Farhadi, A.: YOLO9000: Better, Faster, Stronger. In: IEEE Conference on Computer Vision and Pattern Recognition. pp. 7263-7271 (2017).
6. Farhadi, A., Redmon, J.: Yolov3: An incremental improvement. In: IEEE Conference on Computer Vision and Pattern Recognition. pp. 1–6 (2018).
7. Ge, Z., Liu, S., Wang, F., Li, Z., Sun, J.: YOLOX: Exceeding YOLO Series in 2021. arXiv preprint arXiv: 2107.08430 (2021).
8. Liu, W., Anguelov, D., Erhan, D., Szegedy, C., Reed, S., Fu, C.-Y., Berg, A.C.: SSD: Single Shot MultiBox Detector. In: European Conference on Computer Vision. pp. 21–37 (2016).
9. Kato, S., Brandt, S., Ishikawa, Y., Rajkumar, R.: Operating systems challenges for GPU resource management. In: International Workshop on Operating Systems Platforms for Embedded Real-Time Applications. pp. 23–32 (2011).
10. Bateni, S., Wang, Z., Zhu, Y., Hu, Y., Liu, C.: Co-Optimizing Performance and Memory Footprint Via Integrated CPU/GPU Memory Management, an Implementation on Autonomous Driving Platform. In: 2020 IEEE Real-Time and Embedded Technology and Applications Symposium. pp. 310–323 (2020).
11. Benoit-Cattin, T., Velasco-Montero, D., Fernández-Berni, J.: Impact of thermal throttling on long-term visual inference in a CPU-based edge device. Electronics. 9(12), 2106 (2020).
12. Zhu, X., Xiong, Y., Dai, J., Yuan, L., Wei, Y.: Deep Feature Flow for Video Recognition. In: IEEE Conference on Computer Vision and Pattern Recognition. pp. 2349-2358 (2017).
13. Zhu, X., Wang, Y., Dai, J., Yuan, L., Wei, Y.: Flow-Guided Feature Aggregation for Video Object Detection. In: IEEE International Conference on Computer Vision. pp. 408-417 (2017).
14. Hochreiter, S., Schmidhuber, J.: Long Short-Term Memory. Neural Computation. 9(8), 1735–1780 (1997).
15. Xiao, F., Lee, Y.J.: Video Object Detection with an Aligned Spatial-Temporal Memory. In: European Conference on Computer Vision. pp. 485-501 (2018).
16. SHI, X., Chen, Z., Wang, H., Yeung, D.-Y., Wong, W., WOO, W.: Convolutional LSTM Network: A Machine Learning Approach for Precipitation Nowcasting. Advances in neural information processing systems 28 (2015).

17. Chang, Z., Zhang, X., Wang, S., Ma, S., Ye, Y., Xinguang, X., Gao, W.: MAU: A Motion-Aware Unit for Video Prediction and Beyond. Advances in Neural Information Processing Systems 34, 26950-26962 (2021).

18. Hu, J.-F., Sun, J., Lin, Z., Lai, J.-H., Zeng, W., Zheng, W.-S.: APANet: Auto-Path Aggregation for Future Instance Segmentation Prediction. IEEE Transactions on Pattern Analysis and Machine Intelligence 44(7), 3386–3403 (2022).

19. Mao, H., Kong, T.: CaTDet: Cascaded tracked detector for efficient object detection from video. Proceedings of Machine Learning and Systems 1, 201–211 (2019).

20. Feichtenhofer, C., Pinz, A., Zisserman, A.: Detect to Track and Track to Detect. In: IEEE International Conference on Computer Vision. pp. 3038-3046 (2017).

21. Li, M., Wang, Y.-X., Ramanan, D.: Towards streaming perception. In: European Conference on Computer Vision. pp. 473–488 (2020).

22. Yang, J., Liu, S., Li, Z., Li, X., Sun, J.: Real-time object detection for streaming perception. In: IEEE/CVF Conference on Computer Vision and Pattern Recognition. pp. 5385–5395 (2022).

23. Jo, W., Lee, K., Baik, J., Lee, S., Choi, D., Park, H.: DaDe: Delay-adaptive Detector for Streaming Perception. arXiv preprint arXiv:2212.11558 (2022).

24. Han, Y., Huang, G., Song, S., Yang, L., Wang, H., Wang, Y.: Dynamic Neural Networks: A Survey. IEEE Trans. Pattern Anal. Mach. Intell 44(11), 7436–7456 (2022).

25. Teerapittayanon, S., McDanel, B., Kung, H.T.: BranchyNet: Fast inference via early exiting from deep neural networks. In: International Conference on Pattern Recognition. pp. 2464–2469 (2016).

26. Liu, X., Mou, L., Cui, H., Lu, Z., Song, S.: Finding decision jumps in text classification. Neurocomputing 371, 177–187 (2020).

27. Wang, X., Yu, F., Dou, Z.-Y., Darrell, T., Gonzalez, J.E.: SkipNet: Learning Dynamic Routing in Convolutional Networks. In: European Conference on Computer Vision. pp. 409-424 (2018).

28. Shen, J., Wang, Y., Xu, P., Fu, Y., Wang, Z., Lin, Y.: Fractional Skipping: Towards Finer-Grained Dynamic CNN Inference. In: AAAI Conference on Artificial Intelligence. pp. 5700–5708 (2020).

29. Mullapudi, R.T., Mark, W.R., Shazeer, N., Fatahalian, K.: HydraNets: Specialized Dynamic Architectures for Efficient Inference. In: IEEE Conference on Computer Vision and Pattern Recognition. pp. 8080-8089 (2018).

30. Ehteshami Bejnordi, A., Krestel, R.: Dynamic Channel and Layer Gating in Convolutional Neural Networks. In: Advances in Artificial Intelligence. pp. 33–45 (2020).

31. Li, Y., Song, L., Chen, Y., Li, Z., Zhang, X., Wang, X., Sun, J.: Learning Dynamic Routing for Semantic Segmentation. In: IEEE/CVF Conference on Computer Vision and Pattern Recognition. pp. 8553-8562 (2020).

32. Liu, L., Deng, J.: Dynamic Deep Neural Networks: Optimizing Accuracy-Efficiency Trade-Offs by Selective Execution. AAAI Conference on Artificial Intelligence 32(1) (2018).

33. Lin, T.-Y., Maire, M., Belongie, S., Hays, J., Perona, P., Ramanan, D., Dollár, P., Zitnick, C.L.: Microsoft COCO: Common Objects in Context. In: European Conference on Computer Vision. pp. 740–755 (2014).

34. sAP: Code for Towards Streaming Perception, https://github.com/mtli/sAP, last accessed 2023/6/20.